\documentclass{article}

     \PassOptionsToPackage{numbers, compress}{natbib}

     \usepackage[preprint]{neurips_2025}

\usepackage[utf8]{inputenc} 
\usepackage[T1]{fontenc}    
\usepackage{hyperref}       
\usepackage{url}            
\usepackage{booktabs}       
\usepackage{amsfonts}       
\usepackage{nicefrac}       
\usepackage{microtype}      
\usepackage{xcolor}         

\usepackage{graphicx}
\usepackage{amsmath}
\usepackage{multirow}
\usepackage{algorithm,algorithmic}
\usepackage{caption}

\title{Gradient Propagation in Retrosynthetic Space: \\ An Efficient Framework for Synthesis Plan Generation}

\author{%
  Chengyang Tian \\
  Harbin Institute of Technology \\
  \texttt{tiancy07@qq.com} \\
  \And
  Yuhang Chang \\
  Harbin Institute of Technology \\
  \texttt{2983830542@qq.com} \\ 
  \And
  YangPeng Zhang \\
  Harbin Institute of Technology \\
  \texttt{zhangyp94@163.com} \\
  \And
  Yang Liu\thanks{Corresponding author.} \\
  Harbin Institute of Technology \\
  \texttt{liuyang@hit.edu.cn}
}

\begin{document}

\maketitle

\begin{abstract}
  Retrosynthesis, which aims to identify viable synthetic pathways for target molecules by decomposing them into simpler precursors, is often treated as a search problem. However, its complexity arises from multi-branched tree-structured pathways rather than linear paths. Some algorithms have been successfully applied in this task, but they either overlook the uncertainties inherent in chemical space or face limitations in practical application scenarios. To address these challenges, this paper introduces a novel gradient-propagation-based algorithmic framework for retrosynthetic route exploration. The proposed framework obtains the contributions of different nodes to the target molecule's success probability through gradient propagation and then guides the algorithm to greedily select the node with the highest contribution for expansion, thereby conducting efficient search in the chemical space. Experimental validations demonstrate that our algorithm achieves broad applicability across diverse molecular targets and exhibits superior computational efficiency compared to existing methods.
\end{abstract}

\section{Introduction}

Retrosynthesis, a fundamental task in organic chemistry, is the process of deconstructing a target molecule into a sequence of feasible precursor compounds, ultimately leading to commercially available starting materials. By working backward from the desired product, chemists aim to devise practical and cost-effective synthetic routes. Traditionally, this task relies heavily on expert knowledge and heuristic rules\cite{corey1969}. However, the advent of automatic retrosynthesis algorithms\cite{mcts, star, fallback} has transformed the field, framing the problem as a search task in which the target molecule serves as the initial node, and the search terminates at purchasable building blocks.

An emerging challenge in retrosynthesis algorithms involves the incomplete understanding of reaction mechanisms and unpredictable side reactions, which introduce uncertainties into the results\citep{end2end_retro}. To address this problem, modern algorithms require the capability to model uncertainty and enhance the overall probability of success. More specifically, practical applications demand more than identifying a single optimal route. Generating synthesis plans with multiple routes is essential to maximize the likelihood of successful laboratory execution.

Current retrosynthesis algorithms, particularly retro*\citep{star} and algorithms based on Monte Carlo Tree Search (MCTS)\citep{mcts}, predominantly focus on identifying single optimal synthesis route. While MCTS aims to minimize the search time\cite{kishimoto2019}, retro* tries to find cost-minimized routes using the method derived from A* algorithms. However, neither approach addresses the critical need for generating alternative synthesis routes or optimizing multi-route synthesis plans\cite{heifets2012}. For example, retro* algorithm exhibits such behavior: upon discovering a successful route, the algorithm continues refining this already successful route rather than exploring another backup route. Recent attempts to model the uncertainty in chemical space, such as the retro-fallback algorithm\citep{fallback}, introduced the Successful Synthesis Probability (SSP) metric to evaluate the overall success probabilities across multiple synthesis routes. While the retro-fallback algorithm has succeeded in improving the SSP of synthesis plans, a few limitations remain. First, the stochastic process employed in retro-fallback algorithm incurs substantial computational overhead, particularly when the number of samples scales large. Second, it is difficult for stochastic processes to capture nuances in success possibilities between pathways. Third, while retro-fallback algorithm accelerates computational efficiency by assuming simultaneous expansion of all leaf molecular nodes during node selection process, this simplification conflicts with its actual node expansion behavior, where only a single node is expanded in each step. This misalignment could results in systematic evaluation biases.

In this paper, we present a gradient-based retrosynthesis planning framework that maximizes the Successful Synthesis Probability (SSP) of a synthesis plan. Our primary contribution lies in introducing gradient propagation to quantify individual node influence on overall success probabilities. By establishing connections between leaf nodes and the root node through differentiable formulas, our method dynamically calculates node influence scores that reflect each molecule's potential in improving the SSP. The gradient-derived metric then drives a greedy influence-maximization strategy, where the node with the highest propagated influence score undergoes prioritized expansion at each search iteration. Furthermore, our proposed framework demonstrates significant advantages in both algorithmic generality and computational efficiency.

\section{Related Works}

\paragraph{Problem Statement.} Retrosynthetic planning, the process of identifying viable synthesis pathways for target molecules, has been widely formulated as a search task over directed trees or graphs. Given a target molecule $M_{root}$ and a set of purchasable molecules $\mathcal{I}$, the core objective is to generate synthesis routes and plans: A valid synthesis route for a molecular node $M$ is defined as a tree rooted at $M$, where all leaf nodes are purchasable molecules, and every non-leaf OR node (molecule node) has exactly one child. A synthesis plan for a molecule $M$ may comprise multiple synthesis routes, allowing OR nodes to have multiple successors\cite{rogers2010}.

Existing approaches typically employ a one-step reaction prediction model\cite{one_step_model_template1, one_step_model_template2, one_step_model_logic, one_step_model_lstm, one_step_model_transformer,coley2019} to iteratively decompose molecules into precursors, expanding the search space until purchasable starting materials are reached. Formally, a search graph $\mathcal{G}$ is initialized with a node $M_{root}$ corresponding to the target molecule. In each iteration during the search, a molecular node $M\in\mathcal{G}$ is selected, and a backward model $B(M):= \big(R_i, \mathcal{S}_i\big)_{i=1}^k$ generates $k$ reactions, where $\mathcal{S}_i=\{M_{i1},M_{i2},\cdots\}$ is the set of reactants for reaction $R_i$. Then the graph $\mathcal{G}$ expands with these reactions and molecules\cite{gottipati2020}.

Typically, two types of search structures are used in this task. Algorithms based on MCTS\cite{mcts} adopt an OR tree framework, and other state-of-the-art algorithms, such as retro*\cite{star} and retro-fallback\cite{fallback}, utilize AND-OR tree/graph (see Appendix~\ref{a_structure} for details). This search structure encodes the logical dependencies inherent in chemical reaction space: the successful acquisition of a product molecule (OR node) requires only one valid reaction pathway, whereas the execution of any reaction (AND node) necessitates the availability of all its reactant precursors\cite{chen2021}.

\begin{figure}[ht]
\centering
\includegraphics[height=0.35\textwidth]{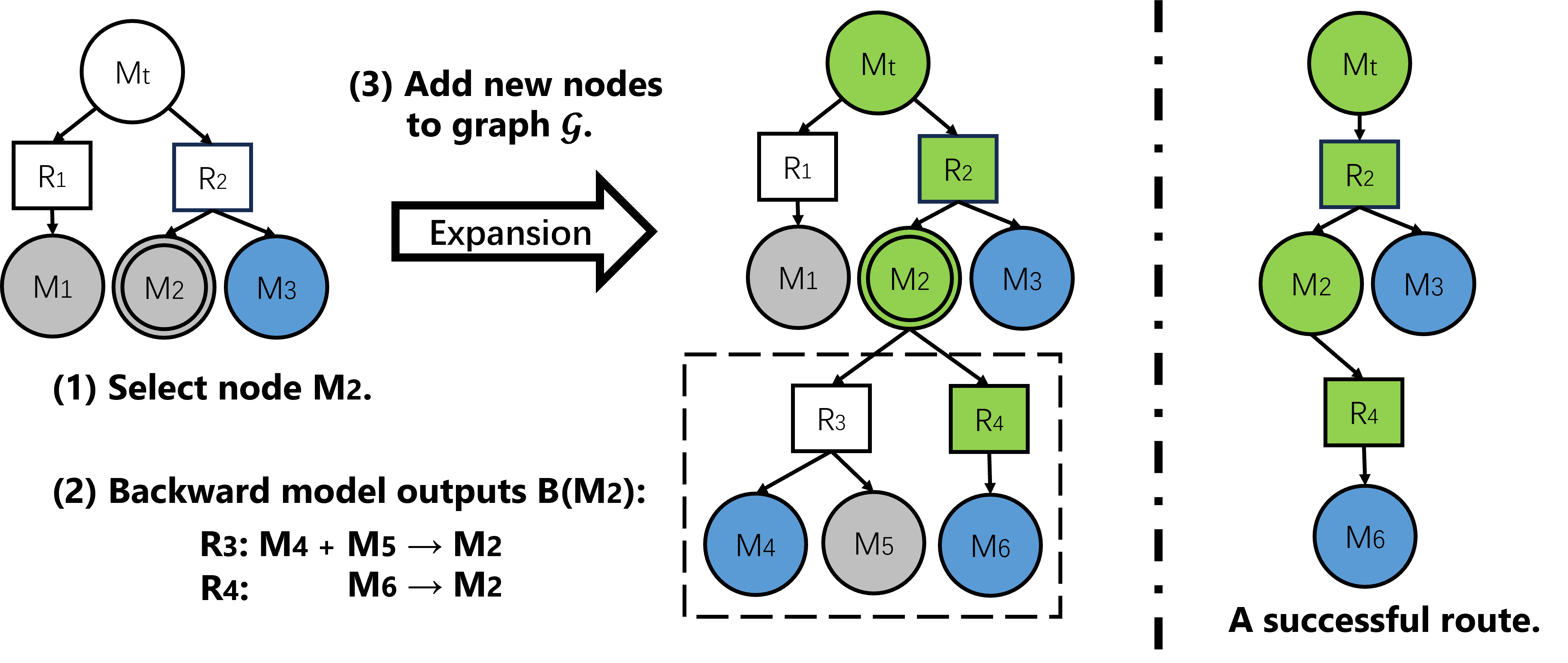}
\caption{The expansion process of an AND-OR tree. Blue nodes represents purchasable molecules, while gray nodes are other unpurchasable leaf nodes. Green nodes are successful nodes within a route. Reaction $R_3$ is NOT viable because it is an AND node with one unsuccessful node $M_5$.}
\label{tree_expansion}
\end{figure}

\paragraph{Retrosynthesis Algorithms.} Retrosynthesis algorithms operate through iterative search processes, each iteration comprising three key phases: \textbf{selection}, \textbf{expansion}, and \textbf{update}. During the selection phase, the algorithm prioritizes molecular nodes in the search graph $\mathcal{G}$ based on predefined heuristics or value estimates. In the expansion phase, the chosen node $M$ undergoes decomposition using the backward model $B(M)$, generating new reaction nodes and molecule nodes. These new nodes are incorporated into $\mathcal{G}$, progressively building potential synthesis routes. Finally, the update phase propagates statistical information (e.g., visit counts, cost estimates) backward through the graph to refine future selection decisions. Figure~\ref{tree_expansion} demonstrates the expansion precess of an AND-OR tree.

Retro*\cite{star} is one of the state-of-the-art retrosynthesis algorithms. The retro* algorithm adopts an A*-inspired approach, defining a node selection rule as:
\begin{equation}\label{star_select_rule}
\begin{aligned}
    &M_{next}\leftarrow\mathop{\arg\min}_M\space v(M) \\
    &\text{where } v(M)=g(M)+h(M)
\end{aligned}
\end{equation}
where $g(M)$ represents the cumulative reaction cost from the root $M_{root}$ to $M$, and $h(M)$ is a heuristic function estimating the remaining synthesis cost from $M$ to purchasable molecules. In practice, $g(M)$ incorporates the reaction cost which is derived from the confidence scores of backward model, while $h(M)$ either employs an optimistic heuristic where $h(M)=0$ or is computed using pre-trained cost predictor.

\paragraph{Modeling Uncertainty in Chemical Space.} While backward models generate candidate reactions for molecular decomposition, their outputs are probabilistic and do not guarantee experimental feasibility. Consequently, synthesis routes generated by algorithms may fail due to invalid reaction steps. To mitigate this risk, modern algorithms will generate synthesis plans with multiple alternative pathways, thereby increasing the likelihood that at least one route is experimentally viable.

The Successful Synthesis Probability (SSP), introduced by retro-fallback\cite{fallback}, quantifies this uncertainty by evaluating the probability that an algorithm’s output contains at least one feasible route. SSP evaluates the final $\mathcal{G}$ generated by algorithms and is defined as:
\begin{equation}\label{ssp_def}
    SSP(\mathcal{G})=\mathbb{P}\left[\exists~Route \subseteq \mathcal{G}, \text{with }\sigma(Route) = 1\right]
\end{equation}
where $\sigma(Route)=1$ means the $Route$ is successful. See Appendix~\ref{a_def_ssp} for detailed definition of SSP. The exact computation of SSP is proven to be NP-hard\cite{fallback, ssp_np}. Retro-fallback employ stochastic processes to approximate SSP, yet such methods incur substantial computational overhead and sometimes fail to discern subtle differences in probability values under specific scenarios.

This paper proposes an efficient algorithmic framework that aims at maximizing the SSP of generated synthesis plans. By employing differentiable SSP estimation, our method uses gradient propagation for node evaluation. Our approach achieves high efficiency while demonstrating broader applicability across diverse scenarios.

\section{Methods}

\subsection{Algorithm Overview}

The algorithm proposed in this paper follows the "Selection-Expansion-Update" iterative workflow commonly adopted by mainstream retrosynthesis algorithms. During the selection phase, the algorithm employs a greedy strategy which aims at selecting the node whose expansion would maximally increase the SSP of the root node:
\begin{equation}\label{selection}
\begin{aligned}
M_{next}=\mathop{\arg\max}\limits_{M\in\mathcal{O}}\Big[SSP\big(\mathcal{G}\big)\big|expand(M)\Big]
\end{aligned}
\end{equation}
where $\big[SSP\big(\mathcal{G}\big)\big|expand(M)\big]$ means the SSP of the search graph $\mathcal{G}$ upon the expansion of molecule node $M$. And $\mathcal{O}$ is the set of "open nodes", which are candidate molecular nodes that can be expanded. Generally, open nodes are unpurchasable leaf nodes that have not yet been expanded. However, not all leaf nodes are open nodes, as some may have been expanded but failed to produce any reactions as their successors.

However, the algorithm does not physically expand any node during the selection phase, so its selection must be determined based on the information acquired from the update phase. In our algorithm, the update phase includes a bottom-up success probability calculation (Section~\ref{bottom_up}) and a top-down gradient propagation (Section~\ref{gradient_propagation}) to assess each node's influence on the target molecule's success probability.

\subsection{Differentiable SSP Estimation}\label{SSP_formulas}

While Equation~\ref{ssp_def} formally defines the Successful Synthesis Probability (SSP), it is not practical for SSP calculation. In retro-fallback's methods, SSP is estimated by employing stochastic processes and sampling-based approximations. In this section, we introduce a differentiable formulation to enable efficient SSP estimation.

Formally, we define the synthesis success probability $\mathbb{P}(X)$ of a node $X$ as the SSP of its corresponding subgraph rooted at $X$. A molecular node is deemed successful if either the molecule is purchasable or at least one of its child reaction nodes is successful. Meanwhile, for a reaction node to be successful, the reaction must be feasible and all of its child nodes (reactant molecules) must also be successful. Therefore, we can summarize the following formula:
\begin{equation}\label{s1}
\begin{aligned}
\mathbb{P}(M)&=\begin{cases}
	1,&m\in\mathcal{I}\\
	\mathbb{P}(R_1\vee R_2\vee R_3\vee\cdots),&\text{otherwise}
	\end{cases}\\
\mathbb{P}(R)&=f(R)\cdot\mathbb{P}(M_1\land M_2\land M_3\land \cdots)
\end{aligned}
\end{equation}
where $m$ is the molecule of node $M$, $\mathcal{I}$ is the set of purchasable molecules, $f(R)$ is the feasible probability of reaction node $R$, $R_i (i=1,2,3,\cdots)$ are successor nodes of $M$, and $M_i (i=1,2,3,\cdots)$ are successor nodes of $R$. This recursive formulation is not well-defined in cyclic graphs (see Appendix~\ref{a_cyclic_graph} for details), we will handle this issues later in Section~\ref{bottom_up}.

Precisely calculating SSP using Equations~\ref{s1} is still impractical as it is a NP problem\cite{fallback, ssp_np}. By assuming independence between the success events of $R_i$ and $R_j$ when $i\neq j$, the following simplified formulas can be used for the probability calculation:
\begin{equation}\label{s_wrong}
\begin{aligned}
\mathbb{P}(M)&=\begin{cases}
	1,&m\in\mathcal{I}\\
	1-\prod_{R\in{Suc(M)}}\left(1-\mathbb{P}(R)\right),&\text{otherwise}
	\end{cases}\\
\mathbb{P}(R)&=f(R)\times\prod_{M\in{Suc(R)}}\mathbb{P}(M)
\end{aligned}
\end{equation}
where $Suc(X)$ represents the set of successors of node $X$ in graph $\mathcal{G}$.

Unfortunately, the success of two different nodes is NOT always independent because them may share some nodes as their children and grandchildren (see Appendix~\ref{a_non_independent} for an example). Actually there are: 
\begin{equation}\label{s2}
\begin{aligned}
\mathbb{P}(R_i\vee R_j) &\le 1-(1-\mathbb{P}(R_i))\cdot(1-\mathbb{P}(R_j)) \\ 
\mathbb{P}(M_i\land M_j) &\ge \mathbb{P}(M_i)\cdot\mathbb{P}(M_j)
\end{aligned}
\end{equation}

In this paper, we combine Equations~\ref{s1} with Inequations~\ref{s2} to calculate an approximation of SSP. Meanwhile, we introduce the parameter $s_0>0$, which is a small positive number. Such setting is meaningful: if the success probabilities of open nodes were set to be zero, it would cause the algorithm to lack awareness of unsuccessful pathways and overlook chemical reactions containing multiple unsuccessful reactants. (Further discussions are in Section~\ref{section_sensitivity} and Appendix~\ref{a_s0}.) Let $s(X)$ denote an approximation of $\mathbb{P}(X)$, we then propose Equations~\ref{s}:
\begin{equation}\label{s}
\begin{aligned}
s(M)&=\begin{cases}
	    1,&m\in\mathcal{I} \\
        s_0, &m\in \mathcal{O} \\ 
	\left[1-\prod_{R\in{Suc(M)}}\left(1-s(R)\right)\right]\times \theta_{\mathcal{G}}(M),&\text{otherwise}
    \end{cases} \\ 
s(R)&=f(R)\times\prod_{M\in{Suc(R)}}s(M)\times \theta_{\mathcal{G}}(R)
\end{aligned}
\end{equation}
where $f(R)$ is the feasible probability of reaction node $R$, $\theta_{\mathcal{G}}(M)$ and $\theta_{\mathcal{G}}(R)$ are coefficient functions which adjust the probabilities for non-independent cases. According to Inequality~\ref{s2} there are $\theta_{\mathcal{G}}(M)\in (0, 1]$ and $\theta_{\mathcal{G}}(R)\ge 1$. In this paper, we view $\theta_{\mathcal{G}}(M)$ and $\theta_{\mathcal{G}}(R)$ as hyperparameters, but they could also be functions derived from the structure of the search graph $\mathcal{G}$.

\subsection{Bottom-up S-Value Calculation}\label{bottom_up}

In acyclic graphs, the s-values of all nodes can be computed by systematically applying Equations~\ref{s} following the (reversed) topological order of nodes. However, in cyclic graphs, the absence of a complete topological order makes Equations~\ref{s} impractical for calculating s-values for all nodes. To address this limitation while enhancing computational efficiency, we propose a bottom-up s-value calculating strategy, as outlined below.

\begin{figure}[ht]
\centering
\includegraphics[width=0.9\textwidth]{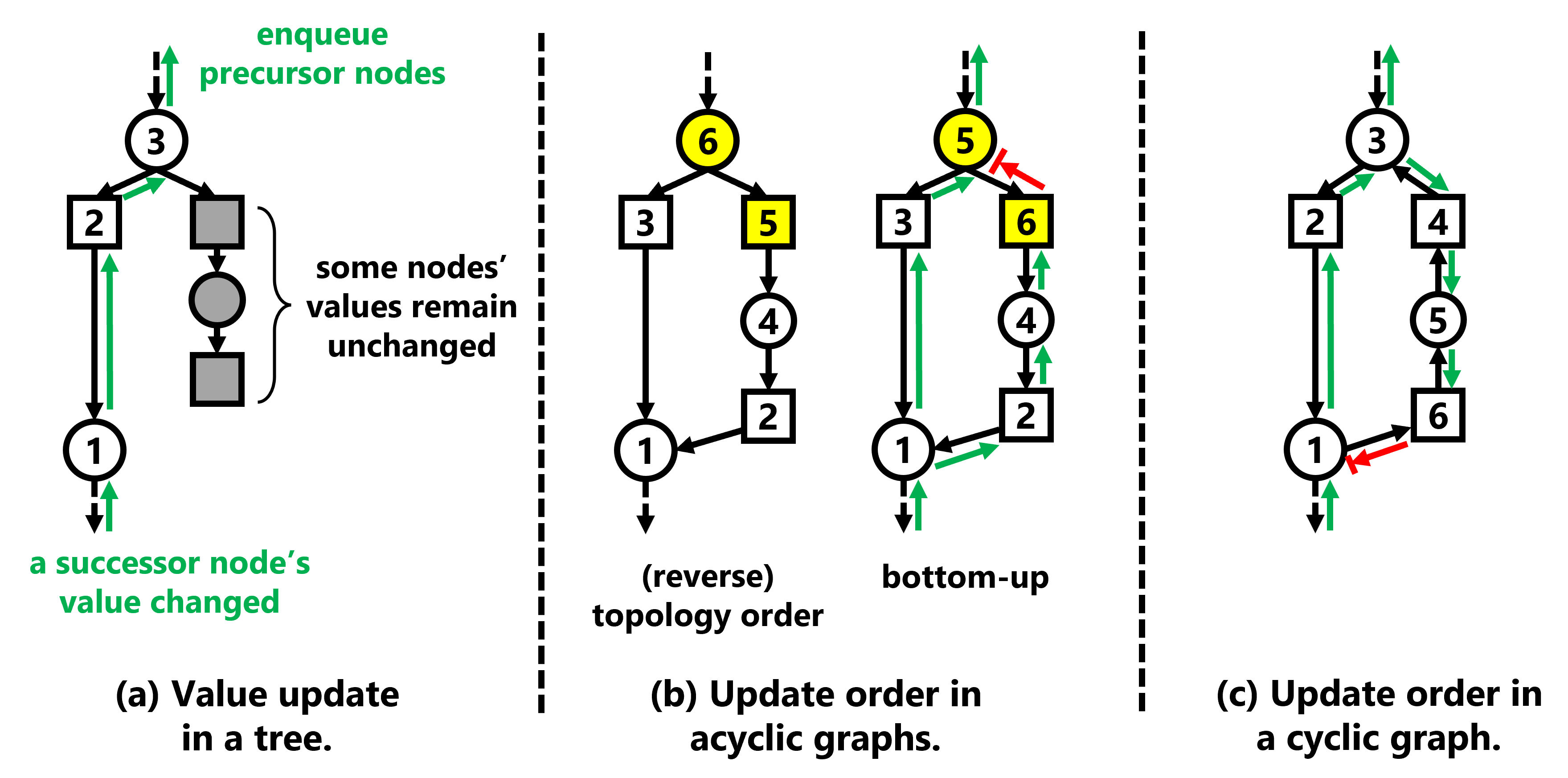}
\caption{Circles represent molecule nodes, and rectangles represent reaction nodes. The numbers denotes the calculation order of each node. (a) In a tree structure, the bottom-up order is equivalent to the reversed topological order. (b) In acyclic graphs, the bottom-up calculation order may differ from the reversed topological order. In the bottom-up update, the top node is only enqueued once. (c) In a cyclic graph, loops are avoided by not to enqueue the node "1" twice.}
\label{information_flow}
\end{figure}

The bottom-up calculation begins by initializing a queue with leaf nodes requiring updates. At each iteration, a node is dequeued from the queue, and its s-value is calculated using Equations~\ref{s}. Then all \textbf{unvisited} (which means that nodes already processed during the current iteration will not be recalculated) predecessors of this node are enqueued. This process repeats until the queue is emptied. The key point is that this strategy ensures that each node is updated only once, thereby eliminating infinite loops in cyclic graphs and avoid the computational overhead associated with topological sorting. Figure~\ref{information_flow} demonstrates how the bottom-up calculation runs in three typical substructures. 

While this bottom-up approach may introduce transient inaccuracies due to deviations from a strict topological order, such errors are non-cumulative. The errors originate from scenarios where a node's value calculation uses outdated data from its child nodes, as its children might be updated later than itself. However, such errors will not accumulate as the algorithm iterates: Subsequent iterations will revisit those affected nodes and recalculate their s-values.

\subsection{Top-down Gradient Propagation}\label{gradient_propagation}

The top-down gradient propagation enables the algorithm to quantify the influence of each node in the search graph on the success probability of the root node (target molecule). Specifically, the success probability of the root node is modeled as a multivariate function, where the variables correspond to the success probabilities of all leaf nodes (i.e., nodes without successors) in the search graph:
\begin{equation}\label{multivariate_function}
\begin{aligned}
    s(M_{root}) = F\Big[s(M_1),s(M_2),\ldots,s(M_k)\Big]
\end{aligned}
\end{equation}
where $M_i(i=1,2,\ldots,k)$ denote leaf nodes, $s(M)$ represents the success probability of node $M$, and $F(\cdot)$ is the multivariate function that aggregates the success probabilities of leaf nodes to estimate the root node’s overall success probability (SSP).

The objective of maximizing the root node’s success probability can be formulated as a optimization problem:
\begin{equation}\label{optimization}
\begin{aligned}
    \text{Maximize}\quad &s(M_{root}) = F\left[s(M_1),s(M_2),\ldots,s(M_k)\right] \\ 
    \text{subject to}\quad &s(M_i) \in (0, 1], \quad i = 1,2,\ldots,k
\end{aligned}
\end{equation}
In this work, the iterative greedy search process mimics coordinate descent optimization\cite{Ruder_2016, Wright_2015} to solve this problem. During each iteration, the algorithm selects the node whose expansion is expected to maximize the incremental in $s(M_{root})$ . This is equivalent to identifying the direction of steepest ascent in the gradient of $F(\cdot)$ with respect to the current node’s success probability and prioritizing the expansion of nodes along this direction. Equation~\ref{derivative} formally describes such node selection strategy.
\begin{equation}\label{derivative}
\begin{aligned}
M_{next}=\mathop{\arg\max}\limits_{M\in\mathcal{O}}\frac{\partial F}{\partial{s(M)}}=\mathop{\arg\max}\limits_{M\in\mathcal{O}}\frac{\partial s(M_{root})}{\partial{s(M)}}
\end{aligned}
\end{equation}

For simplicity, let $D(X)$ denote the derivative $\partial s(M_{root})/\partial s(X)$. By analyzing the structure of the function $F(\cdot)$ (refer to Equation~\ref{s}), the gradient propagation rules are as follows:
\begin{equation}\label{chain}
\begin{aligned}
D(M)&=\begin{cases}
    1,&M\text{ is }M_{root} \\ 
    \sum_{R\in Pre(M)} \Big[D(R)\cdot\partial s(R)/\partial s(M)\Big],&\text{otherwise}
\end{cases}\\
D(R)&=D(P)\cdot\frac{\partial s(P)}{\partial s(R)}\\
\end{aligned}
\end{equation}
where $Pre(M)$ denotes the set of precursors of node $M$, and $P$ is the sole precursor of node $R$ (according to definitions, any reaction node $R$ has exactly one precursor).

Based on Equations~\ref{s}, the derivatives between adjacent nodes are calculated using Equations~\ref{calculate1} and \ref{calculate2}:
\begin{equation}\label{calculate1}
\begin{aligned}
\frac{\partial s(P)}{\partial s(M)}&=\theta_{\mathcal{G}}(P)\times f(P)\times\prod_{\substack{M_i\in Suc(P)\\M_i\neq M}}s(M_i) \\
\end{aligned}
\end{equation}
\begin{equation}\label{calculate2}
\begin{aligned}
\frac{\partial s(P)}{\partial s(R)}&=\theta_{\mathcal{G}}(P)\times\prod_{\substack{R_i\in Suc(P)\\R_i\neq R}}\big[1-s(R_i)\big] \\
\end{aligned}
\end{equation}
Here in Equation~\ref{calculate1}, $P$ represents one of node $M$'s precursors, and in Equation~\ref{calculate2}, $P$ represents the sole precursor of node $R$.

Using the above formulations, the top-down gradient propagation algorithm efficiently computes $D(X)$, the gradient of $s(M_{root})$ with respect to any node $X$ in the search graph. Similar to the bottom-up s-value calculation in Section~\ref{bottom_up}, the gradient propagation does not strictly follow the topological order either. Instead, it employs a recursive approach, ensuring each node is computed only once, thereby maintaining computational efficiency and avoiding infinite loops caused by cyclic graphs.

We state our algorithm explicitly in Algorithm~\ref{alg_2}. In each iteration, the algorithm first selects the node with the maximal gradient $D(\cdot)$ as the expansion target $M_{next}$, then calls the backward model to expand $M_{next}$, adding new nodes to the search graph. Afterwards, it updates the s-values and the gradients of the relevant nodes, preparing for the next iteration.
\begin{algorithm}
\caption{Algorithm framework}\label{alg_2}
\algsetup{indent=3em}
\begin{algorithmic}[1]
\STATE Initialize the root node $M_{root}$
\STATE Set $D(M_{root})=1.0$
\FOR{$i=0$ to $max\_iteration$}
    \STATE \textbf{Selection}: $M_{next}=\mathop{\arg\max}\limits_{M\in\mathcal{O}}\ D(M)$
    \STATE \textbf{Expansion}: Expand $M_{next}$
    \STATE \textbf{Update} $s(\cdot)$
    \STATE \textbf{Update} $D(\cdot)$ for all nodes
\ENDFOR
\RETURN The search graph $\mathcal{G}$
\end{algorithmic}
\end{algorithm}

\section{Experiments}\label{section_experimets}

We establish our experiment based on the popular USPTO\cite{uspto} dataset, which has been widely used for evaluating multi-step retrosynthesis algorithms. We compare our algorithm to MCTS\cite{mcts}, retro*\cite{star}, and retro-fallback\cite{fallback}. Evaluation metrics include SSP, time per iteration, and memory consumption during the search. The test molecules are 190 "hard" molecules from retro*\cite{star} and 1000 molecules from GuacaMol\cite{guacamol}, which are the same as those reported in retro-fallback's experiments\cite{fallback}.

\subsection{Parameter Sensitivity Analysis: Effects on Algorithm Performances}\label{section_sensitivity}

$s_0$ is a critical parameter in retrosynthetic algorithms that represents the default value assigned to unpurchasable leaf nodes. These nodes, which remain unexplored and unsuccessful in the search graph, require predefined cost/success probability estimates. In retro*, the heuristic function $h$ is used for such cost estimates, while retro-fallback employs a buyability model to determine default success probabilities for molecular nodes.

\begin{figure}[ht]
\centering
\includegraphics[height=0.45\textwidth]{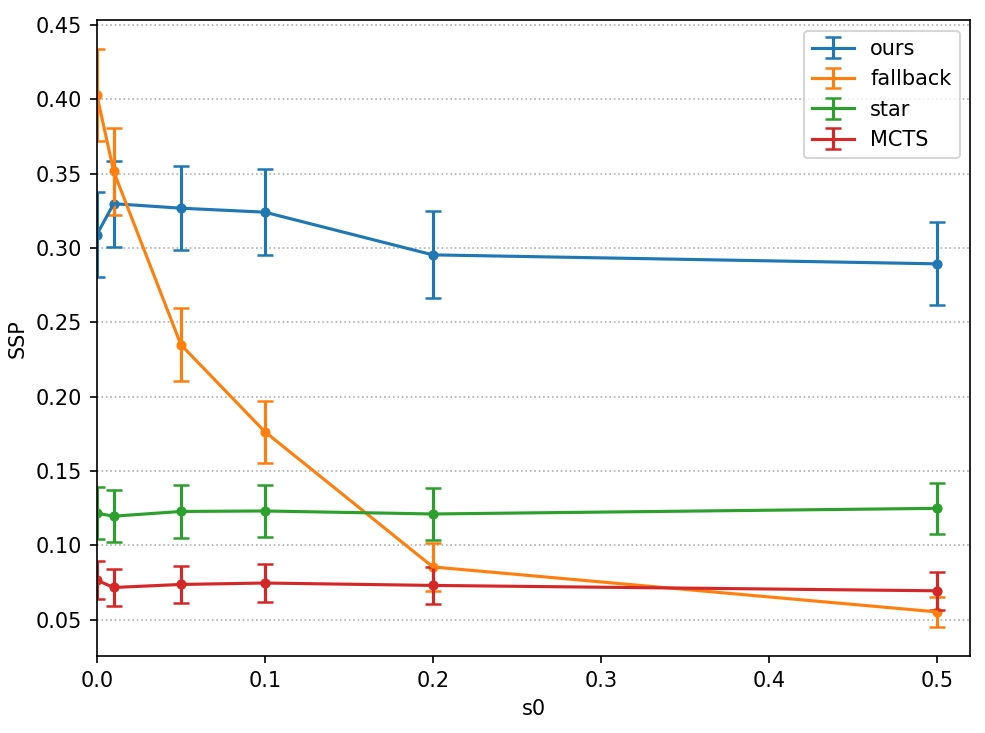}
\caption{Average SSP achieved by algorithms on the 190 "hard" molecules under different default values $s_0$. Error bars represent standard errors. Detail settings are in Appendix~\ref{a_setup}.}
\label{exp_s0}
\end{figure}

To assess the parameter sensitivity concerning $s_0$, we systematically evaluate different algorithms under varying configurations of default success probabilities for unsuccessful leaf nodes. A group of distinct default values $s_0\in\{0.00, 0.01, 0.05, 0.10, 0.20, 0.50\}$ are implemented during the search process to quantify their impacts on algorithmic performance. The experiment results are shown in Figure~\ref{exp_s0}.

The experimental results demonstrate that only when the default value $s_0$ is set to $0$ or $0.01$, retro-fallback achieves slightly higher SSP results than our algorithm. However, as $s_0$ increases further, its performance degrades significantly. Except for the retro-fallback algorithm, other algorithms exhibit insensitivity to parameter $s_0$. This parameter sensitivity of retro-fallback may hinder its applicability in some practical scenarios. For instance, assigning a higher $s_0$ value to a node can encourage the algorithm to explore its brother nodes, while simultaneously increasing the selection frequency of chemical reactions with multiple reactants. The effect of $s_0$ differs from the heuristic function in the retro-fallback method: a higher heuristic score of a node increases the chance to expand itself rather than its brothers. More interpretation of $s_0$'s function is provided in Appendix~\ref{a_s0}.


\subsection{Algorithm Performances on SSP Maximization}

Setting the default success probability $s_0=0.05$ for unpurchasable leaf molecule nodes, we report the averaged results on the two sets of test molecules in Table~\ref{result_190} and Table~\ref{result_1000}, respectively. The results show that our algorithm outperforms MCTS\cite{mcts}, retro*\cite{star} and retro-fallback\cite{fallback}, achieving the highest Successful Synthesis Probability (SSP) with the lowest time per iteration and memory consumption.

\begin{table}[ht]
  \caption{Results of 190 "hard" molecules from retro*.}
  \label{result_190}
  \centering
  \begin{tabular}{c|ccc|ccc}
    \toprule
    \multirow {2}{*}{Algorithm} & \multicolumn{3}{c|}{Constant Feasibility} & \multicolumn{3}{c}{Rank Feasibility}\\ \cline{2-4} \cline{5-7}
     & SSP (\%) & Time (s/iter) & Mem (MB) & SSP (\%) & Time (s/iter) & Mem (MB) \\
    \midrule
    MCTS & 7.06 & 0.73 & 29.8 & 25.3 & 0.84 &77.4\\
    retro* & 12.60 & 0.71 & 231.2 & 24.6 & 0.73 & 284.4\\
    fallback & 24.82 & 0.77 & 99.2 & 30.3 & 0.80 & 93.5 \\
    ours & \textbf{34.37} & \textbf{0.61} & \textbf{14.6} & \textbf{42.0} & \textbf{0.66} & \textbf{14.3}\\
    \bottomrule
  \end{tabular}
\end{table}

\begin{table}[ht]
  \caption{Results of 1000 test molecules from GuacaMol.}
  \label{result_1000}
  \centering
  \begin{tabular}{c|ccc|ccc}
    \toprule
    \multirow {2}{*}{Algorithm} & \multicolumn{3}{c|}{Constant Feasibility} & \multicolumn{3}{c}{Rank Feasibility}\\ \cline{2-4} \cline{5-7}
     & SSP (\%) & Time (s/iter) & Mem (MB) & SSP (\%) & Time (s/iter) & Mem (MB) \\
    \midrule
    MCTS & 51.34 & 0.72 & 30.0 & 69.3 & 0.84 & 74.2 \\
    retro* & 58.12 & 0.69 & 225.2 & 69.0 & 0.68 & 290.2  \\
    fallback & 65.61 & 0.75 & 100.5 & 70.45 & 0.76 & 95.5 \\
    ours & \textbf{72.94} & \textbf{0.62} & \textbf{15.1} & \textbf{74.90} & \textbf{0.61} & \textbf{15.3} \\
    \bottomrule
  \end{tabular}
\end{table}

From the results in Table~\ref{result_190} and Table~\ref{result_1000}, we can also observe that: (1) retro* has significantly the maximal space cost. This is because retro* is tree-based, meaning that there can be multiple nodes for a single molecule. Meanwhile our method and retro-fallback are graph-based, eliminating the cost of redundant nodes, and MCTS algorithm uses OR tree (see its definition in Appendix~\ref{a_structure}) instead of AND-OR tree. (2) retro-fallback consumes more memory than our method. This is probably due to the heavy burden of stochastic processes adopted by retro-fallback. (3) it can be observed that the SSP results of retro-fallback are lower than those reported in their original paper\cite{fallback}. From Section~\ref{section_sensitivity}, we know that this is caused by the different setting of $s_0$. In retro-fallback\cite{fallback}, they adopt a "binary" buyability model which assigns the success probabilities of unsuccessful leaf nodes to zero, while here they are set to $s_0=0.05$.

\subsection{Ablation Studies}

In Section~\ref{abla_bottom_up}, we conduct ablation studies to demonstrate that our bottom-up s-value calculation approach is more efficient compared to the (reversed) topological order calculation. And in Section~\ref{abla_grad}, experimental results validate that our gradient-based methodology significantly improves algorithmic efficiency.

\begin{table}[ht]
  \caption{Ablation results based on 190 "hard" molecules. w/ Topo means the 
s-value calculation follows the reserved topological order instead of using our bottom-up calculation strategy. w/o Grad means the algorithm runs without our gradient propagation strategy.}
  \label{result_ablation}
  \centering
  \begin{tabular}{c|ccc|ccc}
    \toprule
    \multirow {2}{*}{Algorithm} & \multicolumn{3}{c|}{Constant Feasibility} & \multicolumn{3}{c}{Rank Feasibility}\\ \cline{2-4} \cline{5-7}
     & SSP (\%) & Time (s/iter) & Mem (MB) & SSP (\%) & Time (s/iter) & Mem (MB) \\
    \midrule
    original & \textbf{34.37} & \textbf{0.61} & \textbf{14.6} & \textbf{42.0} & \textbf{0.66} & \textbf{14.3} \\
    w/ Topo & 26.76 & 0.87 & 95.4 & 33.14 & 0.92 & 96.4 \\
    w/o Grad & 25.12 & 47.24 & 90.1 & 32.27 & 54.30 & 85.1 \\
    \bottomrule
  \end{tabular}
\end{table}

\subsubsection{Effect of the Bottom-up \& Top-down Calculation Strategy}\label{abla_bottom_up}

As shown in Table~\ref{result_ablation}, the algorithm without bottom-up s-value calculation strategy faces a performance decline in SSP and time consumption. The drop in SSP is probably caused by the deletion of some edges: To obtain the topological order, the search graph must be acyclic, so some edges have to be removed in a cycle. Meanwhile the search time also increases because the topological sorting process of a graph is expensive.

\subsubsection{Effect of the Gradient-based Method}\label{abla_grad}

Our gradient-based approach quantifies each node's influence on target molecule success probability. In contrast, ablation experiments employ a more naive method directly based on Equation~\ref{selection}: By optimistically assuming expanding a node $expand(M)$ will result in $s(M)=1$, we then observe which node's expansion leads to the highest improvement of the root node's SSP, and finally select it as the next expanding node $M_{next}$.

As the results in Table~\ref{result_ablation} show, the use of gradient is of vital important for efficiency. Calculating $s(M_{t}|expand(M))$ for every leaf node makes the algorithm $70\times$ slower.

\section{Discussions and Limitations}\label{limitations}

This study introduces a novel gradient-based search algorithm for multi-step retrosynthetic planning, which aims at maximizing the Successful Synthesis Probability (SSP) to generate high-quality synthesis plans with multiple routes. Experiments demonstrate that our method achieves SSP performance comparable to or exceeding state-of-the-art approaches while maintaining computational efficiency, and is fitted in scenarios with diverse parameter configurations. Beyond retrosynthesis, the proposed gradient-based framework establishes a generalized methodology with broader applicability. The inherent compatibility of the AND-OR graph search structure with formal verification systems, such as those in automated theorem proving, suggests potential for cross-domain adaptation. We hope that this gradient-based paradigm may inspire research in related fields.

This work has limitations that warrant further investigation: (1) Coefficient functions $\theta_{\mathcal{G}}(R)$ and $\theta_{\mathcal{G}}(M)$ require systematic redesign. For instance, these functions may be realized via neural networks, which analyze the search graph's structure and calibrate probability estimations. (2) The use of heuristic functions has not yet been incorporated. Heuristic functions might further improve the search efficiency and solution quality of the algorithm.

Moving forward, we will continue to advance our work to address chemical synthesis challenges. The chemical space abounds with unknowns and uncertainties – we envision that increasingly intelligent machine learning methods will empower researchers to systematically explore the expanding frontiers of molecular and reaction space.

\bibliography{ref}


\newpage
\appendix

\section{Search Structures}\label{a_structure}

\begin{figure}[ht]
\centering
\includegraphics[height=0.35\textwidth]{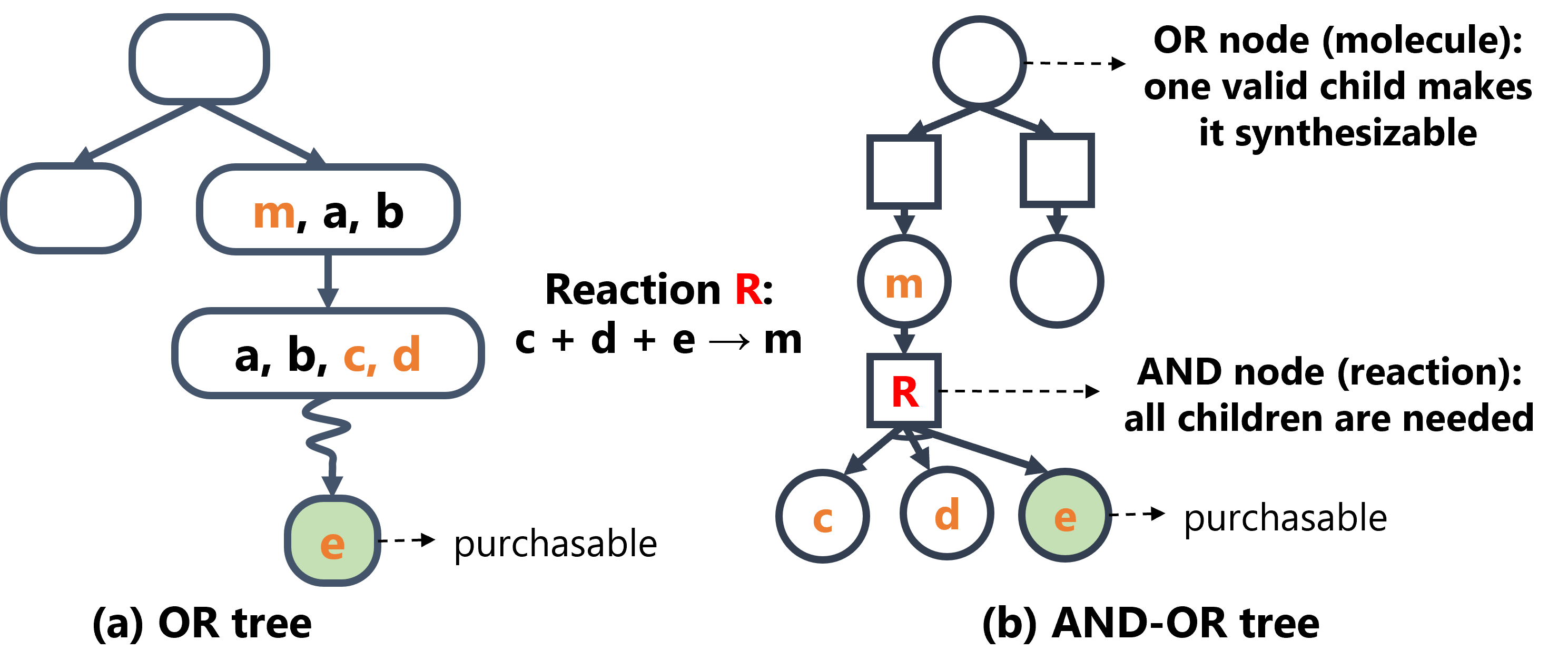}
\caption{The structures of OR tree, AND-OR tree. In an OR tree, each node represents a set of molecules. In an AND-OR tree, there are two types of nodes. Each OR node (circle) represents a single molecule, while each AND node (rectangle) represent a chemical reaction.}
\label{search_structure}
\end{figure}

\paragraph{OR tree.} The OR tree search structure is a fundamental framework adopted in MCTS-based retrosynthesis algorithms\cite{mcts}. In an OR tree, each node $U$ represents a set of molecules $\mathcal{M}_U$, and each of its child node $V\in ch(U)$ comes from running the backward model over a molecule within $U$. For example in Figure~\ref{search_structure} (a), there is an OR node $U:=\{m, a, b\}$, and we run backward model over molecule $m$ to obtain a chemical reaction $R:\ c+d+e\rightarrow m$, where molecule $e$ is purchasable. To represent this, we create a child node $V:=\{a,b,c,d\}$ and associate it with a building block $e$.

\paragraph{AND-OR tree / graph.} An AND-OR graph is a directed graph that capable of capturing the relationships between chemical reactions and molecules. It contains two types of nodes: AND-OR node and OR node. Each OR node represents a single molecule, while each AND node represents a single-product reaction. A chemical reaction with multiple products can be represented by multiple single-product reactions, where each product of the reaction corresponds to one single-product reaction. Edges in an AND-OR graph represent the relationships between reactions and molecules: An edge $M\rightarrow R$ from a molecule node to a reaction node indicates that $M$ is the product of reaction $R$, while an edge $R\rightarrow M$ from a reaction node to a molecule node denotes that $M$ is one of the reactants of reaction $R$. Note that the direction of the arrows in the AND-OR graph is opposite to the direction in which the chemical reaction proceeds. 

These two distinct types of nodes clearly capture the dependency relationships between chemical reactions and molecules: producing a molecule can be achieved through any chemical reaction which has it as a product, while successfully carrying out a chemical reaction depends on the presence of all its reactant molecules.

\subsection{Issues of Cyclic Graphs}\label{a_cyclic_graph}

In directed graphs formed by molecular interactions and chemical reactions, cyclic structures may emerge, such as those created by reversible reactions. When encountering such cycles, Equations~\ref{s1} become ambiguous. Consider a minimal cycle comprising two reciprocal reactions: $R_1:M_1\rightarrow M_2$ and $R_2:M2\rightarrow M_1$. The recursive format of Equation~\ref{s1} leads to infinite regression when trying to calculate the success probability of $M_1$, as it creates a circular dependency chain $M_1\rightarrow R_1\rightarrow M_2\rightarrow R_2\rightarrow M_1\rightarrow\cdots$: $M_1$'s success probability depends on $M_2$'s through $R_1$, while $M_2$'s in turn depends on $M_1$'s through $R_2$.

\begin{figure}[ht]
\centering
\includegraphics[width=0.95\textwidth]{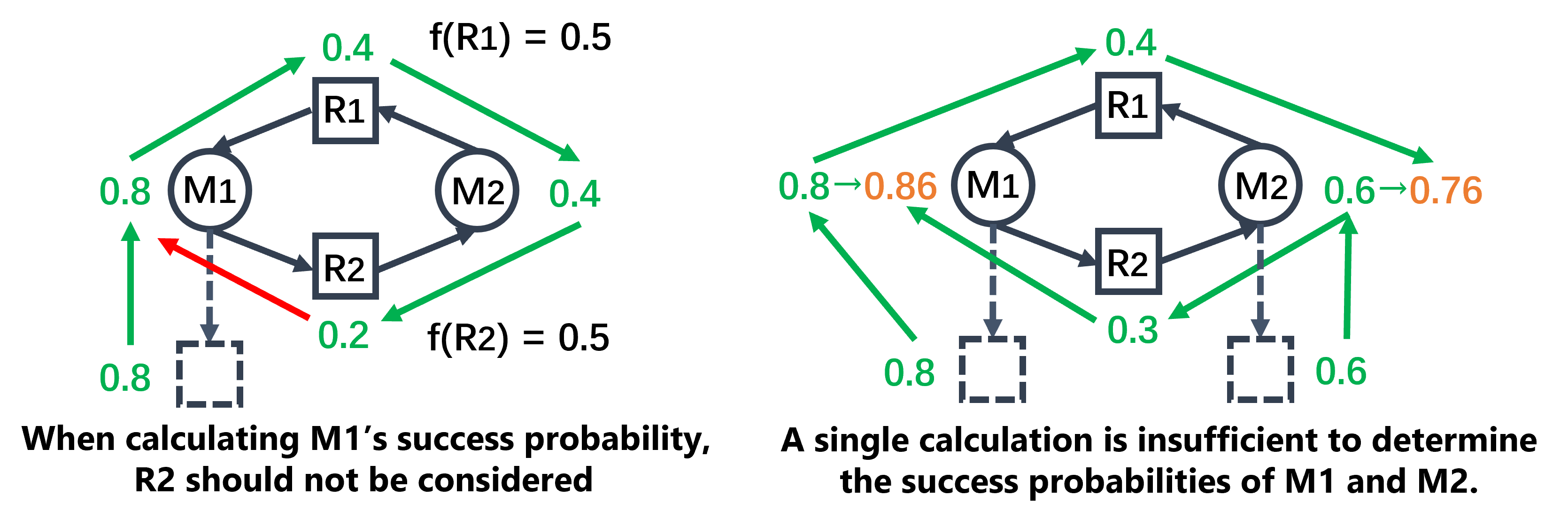}
\caption{Accurate probability calculating process under two cyclic graphs. Green arrows represent the calculating order.}
\label{cycle_calculate}
\end{figure}

Figure~\ref{cycle_calculate} demonstrates the process of correctly calculating probabilities under the two simplest cyclic scenarios. Assuming both reactions' feasibility $f(R_1)=f(R_2)=0.5$. 

In the figure on the left, node $M_1$ has a child reaction node with a success probability of $0.8$, so its own success probability is also $0.8$. Consequently, the success probability of reaction $R_1$ is calculated as $f(R_1)\times0.8 = 0.4$, making the success probability of molecule $M_2$ synthesized via $R_1$ is also $0.4$. The success probability of reaction $R_2$ then becomes $f(R_2)\times0.4 = 0.2$. Although $M_1$ can be synthesized through $R_2$, the existence of $R_2$ does not contribute to $M_1$’s success probability.

In the figure on the right, $M_1$ has a child reaction node with a success probability of $0.8$, while $M_2$ also has a child reaction node with a success probability of $0.6$. Here, we first get the success probabilities of $M_1$ and $M_2$ as $0.8$ and $0.6$, respectively. Next, we calculate the success probability of $R_1$ as $f(R_1)\times0.8 = 0.4$ and that of $R_2$ as $f(R_2)\times0.6 = 0.3$. In this case, $R_1$ and $R_2$ will contribute to the success probabilities of $M_2$ and $M_1$, respectively. Thus, we need to recalculate the success probabilities of these two molecular nodes: $\mathbb{P}(M_1)=1-(1-0.8)\times(1-0.3)=0.86$ and $\mathbb{P}(M_2)=1-(1-0.6)\times(1-0.4)=0.76$.

The presented examples in Figure~\ref{cycle_calculate} demonstrate that precise computation of success probabilities in cyclic structures becomes difficult. In practice, it is even more complicated due to larger cycles and interconnected cyclic topologies. Under such conditions, Equations~\ref{s1} do not serve well for precise probability calculations. To address this issue, we implements an approximate method which follows the "single calculation" principle (in Section~\ref{bottom_up}): each node's probability undergoes no more than one computation during each iteration. This prevents our algorithm from falling into infinite loops.

\section{The Definition of SSP}\label{a_def_ssp}

Successful Synthesis Probability (SSP) is an evaluation metric that quantifies the likelihood that at least one synthesis route in a set of routes will be successful. 

Formally, SSP is defined on the basis of stochastic process. First of all, function $f(R):\mathcal{R}\rightarrow\{0, 1\}$ denotes whether a reaction $R$ is feasible ($f(R)=1$) or not ($f(R)=0$), and function $b(M):\mathcal{M}\rightarrow\{0, 1\}$ denotes whether a molecule $M$ is purchasable ($b(M)=1$) or not ($b(M)=0$). Then, feasibility model $\xi_f$ and $\xi_b$ is defined to be binary stochastic processes over the distribution of $f$ and $b$. Based on $f$ and $b$, $\sigma(Route)$ is defined to represent whether a $Route$ is successful ($\sigma(Route)=1$) or not ($\sigma(Route)=0$):
\begin{equation}
\begin{aligned}
\sigma(Route;f,b)&=\begin{cases}
    1, &f(R)=1\ \forall R\in Route,\text{and }b(M)=1\ \forall M\in\mathcal{F}(Route)\\ 
    0, &\text{otherwise}
\end{cases}\\
\end{aligned}
\end{equation}
where $\mathcal{F}(Route)$ denotes the set of leaf nodes of the tree-like $Route$.

Finally, given a search graph $\mathcal{G}$ containing a set of synthesis routes, its SSP is defined as:
\begin{equation}
    SSP(\mathcal{G}; \xi_f, \xi_b) = \mathbb{P}_{f \sim \xi_f, b \sim \xi_b}\left[ \exists Route\subset\mathcal{G},\text{ with }\sigma(Route; f, b) = 1 \right]
\end{equation}

When graph $\mathcal{G}$ degenerates into a $Route$, $SSP(Route; \xi_f, \xi_b)=\mathbb{E}_{f,b}\left[\sigma(Route;f,b)\right]$ represents the probability that $Route$ is successful.

\section{Non-independent Probabilities Between Nodes}\label{a_non_independent}

\begin{figure}[ht]
\centering
\includegraphics[height=0.3\textwidth]{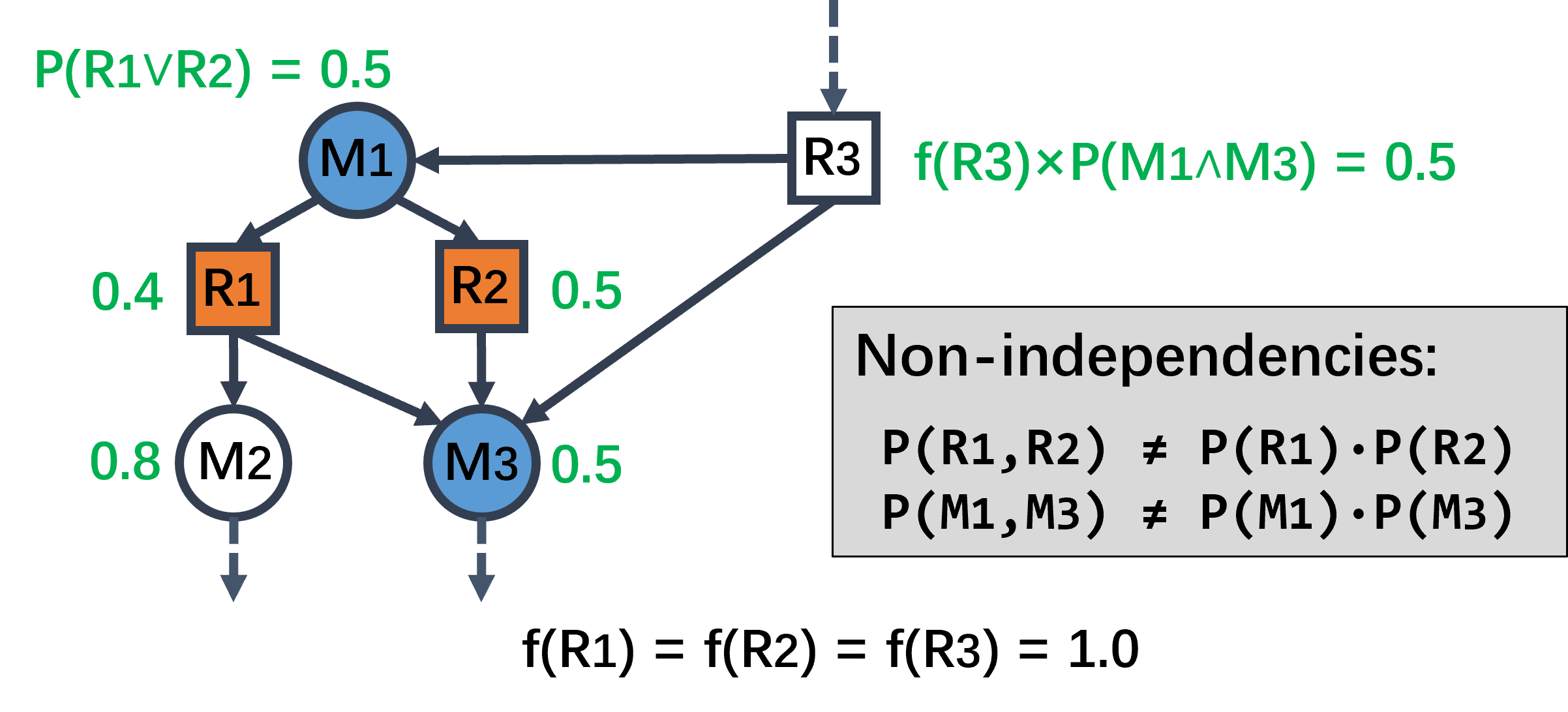}
\caption{An example of non-independency between nodes. Assume the feasible probabilities of all reactions are $1.0$. Numbers besides nodes denote their success probabilities. The success of $R_1$ and $R_2$ is non-independent, as well as node $M_1$ and $M_3$.}
\label{non_ind}
\end{figure}

The success probabilities of two nodes in a search graph may not be independent, as these nodes could share some descendant nodes. This non-independency is demonstrated with an example in Figure~\ref{non_ind}.

Figure~\ref{non_ind} demonstrates the two types of non-independencies. First, reaction nodes $R_1$ and $R_2$ are non-independent as they both rely on $M_3$. In fact, if $R_1$ succeeds, $R_2$ must necessarily succeed. Thus, the success probability of $M_1$, defined as $\text{P}(M_1)=\text{P}(R_1\vee R_2)$, equals $\text{P}(R_2)=0.5$, rather than $1 - (1 - \text{P}(R_1))\times(1 - \text{P}(R_2))=0.7$. Second, $M_1$ and $M_3$ are also non-independent, with $M_1$ succeeding if and only if $M_3$ succeeds. Consequently, the success probability of $R3$ is calculated as $\text{P}(R_3)=f(R_3)\times \text{P}(M_1\land M_3) = 1\times \text{P}(M_3) = 0.5$, which differs from $f(R_3)\times \text{P}(M_1)\times \text{P}(M_3)=0.25$.

\section{Impact of Parameter s0 on Algorithmic Behavior}\label{a_s0}

\begin{figure}[ht]
\centering
\includegraphics[width=0.9\textwidth]{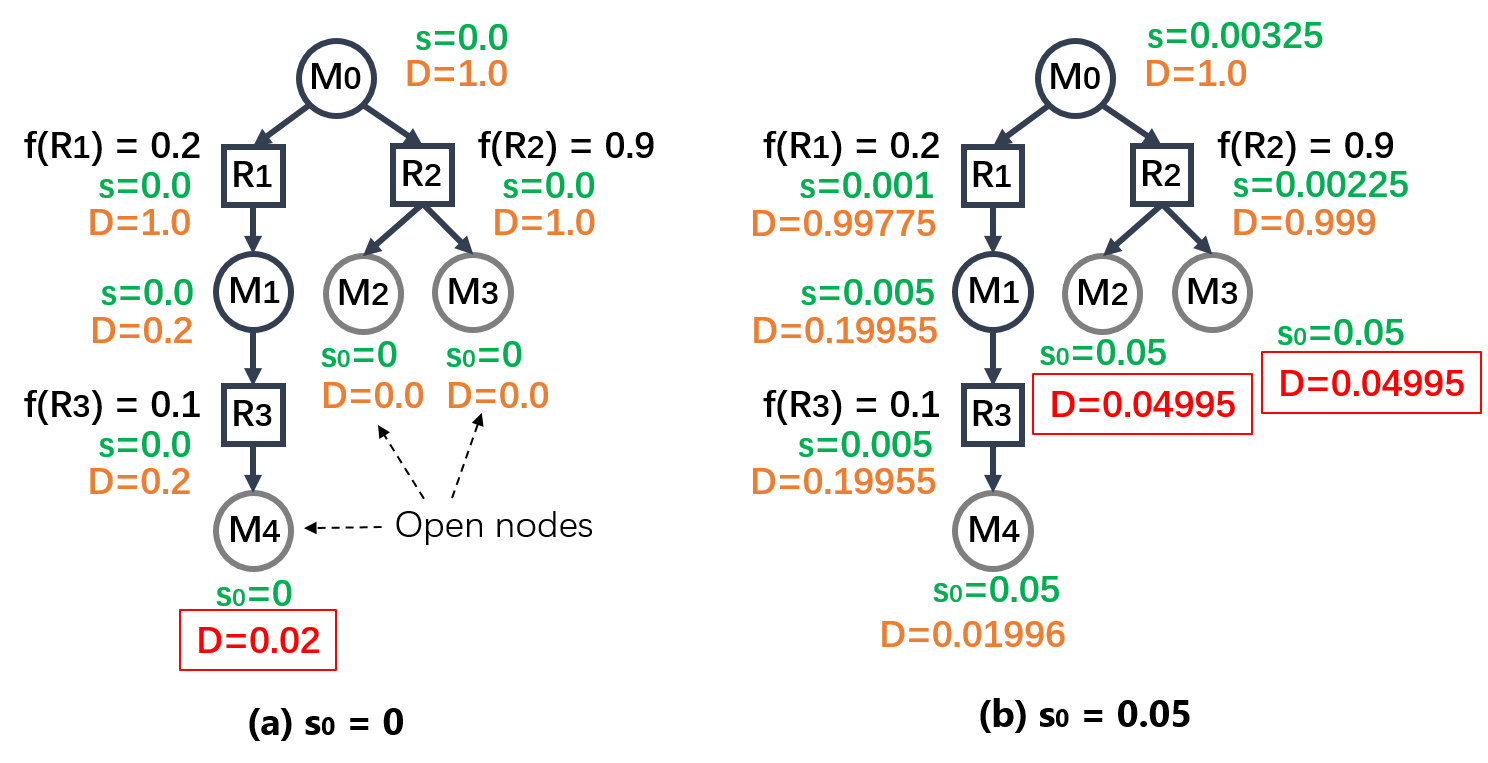}
\caption{The parameter $s_0$ influences algorithm's behavior. The s-value and gradient of each node are annotated as $s$ and $D$, respectively. (a) Setting $s_0=0$, molecule $M_4$ will be chosen as the next expanding node. (b) Setting $s_0=0.05$, the next expanding node will choose between $M_2$ and $M_3$.}
\label{use_of_s0}
\end{figure}

Within the proposed retrosynthesis framework, parameter $s_0$ serves as a tunable control that influences the algorithm's behavioral preferences. Increasing the value of $s_0$ directs the algorithm to prioritize exploration of chemical reactions which have multiple unsuccessful reactants. Figure~\ref{use_of_s0} provides an example.

In Figure~\ref{use_of_s0} (a), when the default success probabilities of $M_2$ and $M_3$ are set to $s_0=0$, their gradient values are always zero according to Equation~\ref{calculate1}, causing the algorithm to overlook them. Actually, if $s_0$ is set to zero, all reactions with multiple unsuccessful reactants will be completely ignored. Meanwhile in Figure~\ref{use_of_s0} (b), setting $s_0$ to a small positive value $0.05$ changes the algorithm's behavior, elevating the priorities of $M_2$ and $M_3$ above $M_4$. This behavior is accordance with our hope since reaction $R_2$ demonstrates a feasible probability $f(R_2)=0.9$, significantly exceeding $f(R_1)=0.2$ and $f(R_3)=0.1$. The algorithm thus reasonably prioritizes exploring $R_2$'s pathways despite its greater number of unsuccessful child nodes.

\section{Details of Experimental Setup}\label{a_setup}

\paragraph{Backward Model.} We adopt the pretrained one-step backward model from retro*\cite{star}, and the top-50 templates selected by the model are considered during the search.

\paragraph{Feasibility Models.} The feasibility model, which determines the feasible probability of each reaction, is basically from retro-fallback\cite{fallback}. The constant feasibility model assigns the same constant $f(R)=0.5$ to every reaction node. And the rank feasibility model provides higher probabilities for reactions that rank high in the backward model's output:
\begin{equation}
f(R)=\frac{0.75}{1+rank/10}
\end{equation}
where $rank=1,2,\ldots$ is the rank of reaction $R$, which comes from the score assigned to each predicted reaction by backward model.

\paragraph{Purchasable Molecules.} The purchasable molecules that serve as the search destination are extracted from the inventory of eMolecules, which is a chemical supplier and provides a list of commercially available molecules. Following the "binary buyability model" in retro-fallback\cite{fallback}, we convert the top-3 tier molecules in eMolecules (which are the most easier to obtain) into our purchasable molecule inventory $\mathcal{I}$.

\paragraph{Others.} Experiments were conducted on an Intel Xeon Silver 4214 CPU and an NVIDIA RTX A6000 GPU. Retro* and retro-fallback use their optimistic heuristics during search. Max expansion depth is set to $30$ for MCTS, while $999$ (functions as $\infty$) for other algorithms.

\section{Demonstration of Experimental Examples}

\begin{figure}[ht]
\centering
\includegraphics[height=1.5\textwidth]{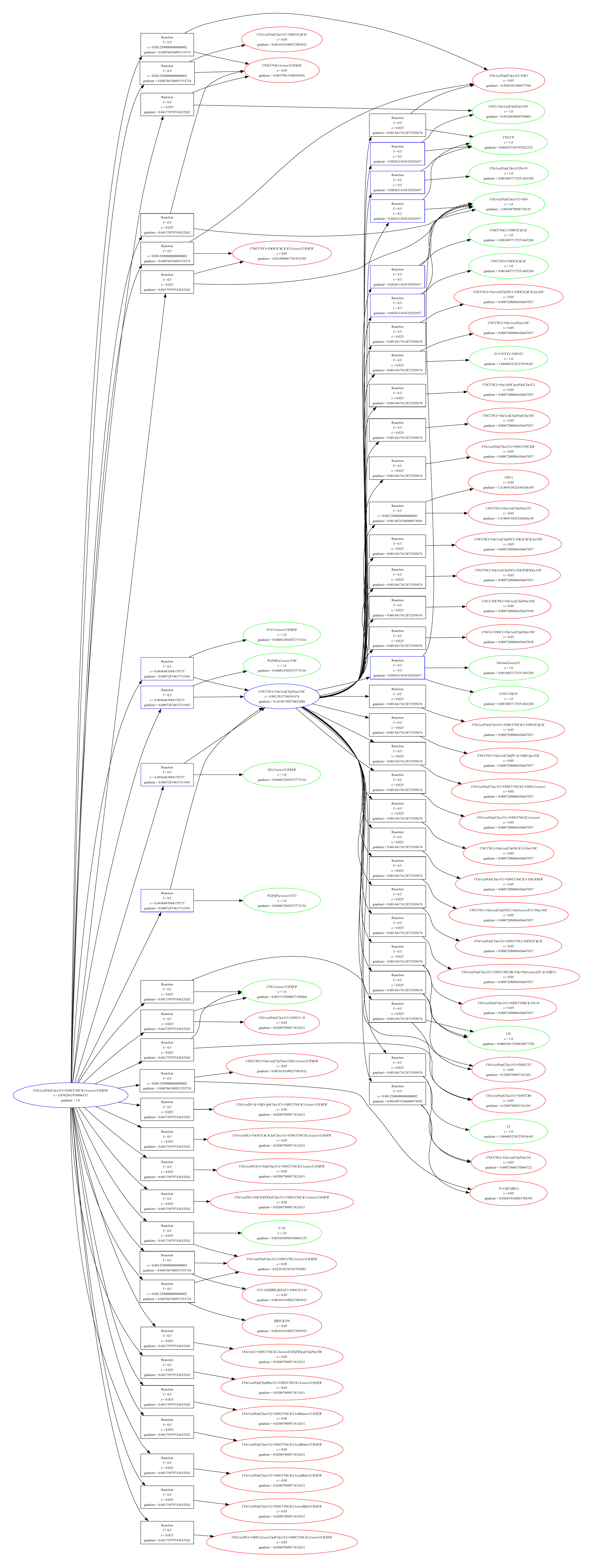}
\caption{The search graph of molecule "COc1ccc(-c2cc3cc(OC)c(OC)cc3c(C)n2)cc1C" at the 2nd iteration. Green nodes denote purchasable molecules, and blue nodes are part of a successful route. Each node's s-value and gradient are annotated in this figure.}
\label{full_graph}
\end{figure}

\begin{figure}[ht]
\centering
\includegraphics[height=1.5\textwidth]{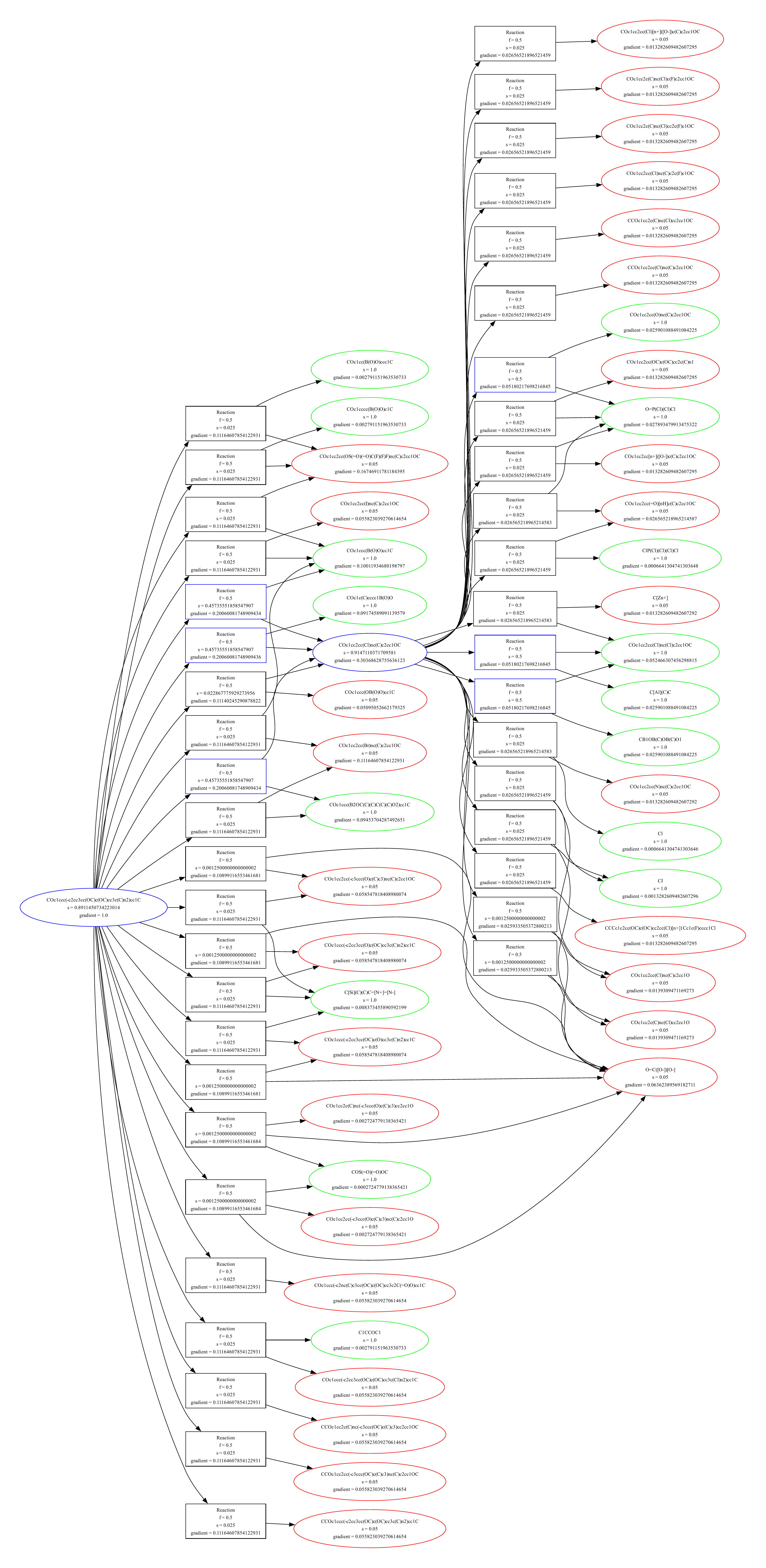}
\caption{The search graph of molecule "COc1cc(N)c(Cl)cc1C(=O)NCCN(C)Cc1ccccc1C(F)(F)F" at the 2nd iteration. Green nodes denote purchasable molecules, and blue nodes are part of a successful route. Each node's s-value and gradient are annotated in this figure.}
\label{full_graph_2}
\end{figure}

\begin{figure}[ht]
\centering
\includegraphics[width=1.0\textwidth]{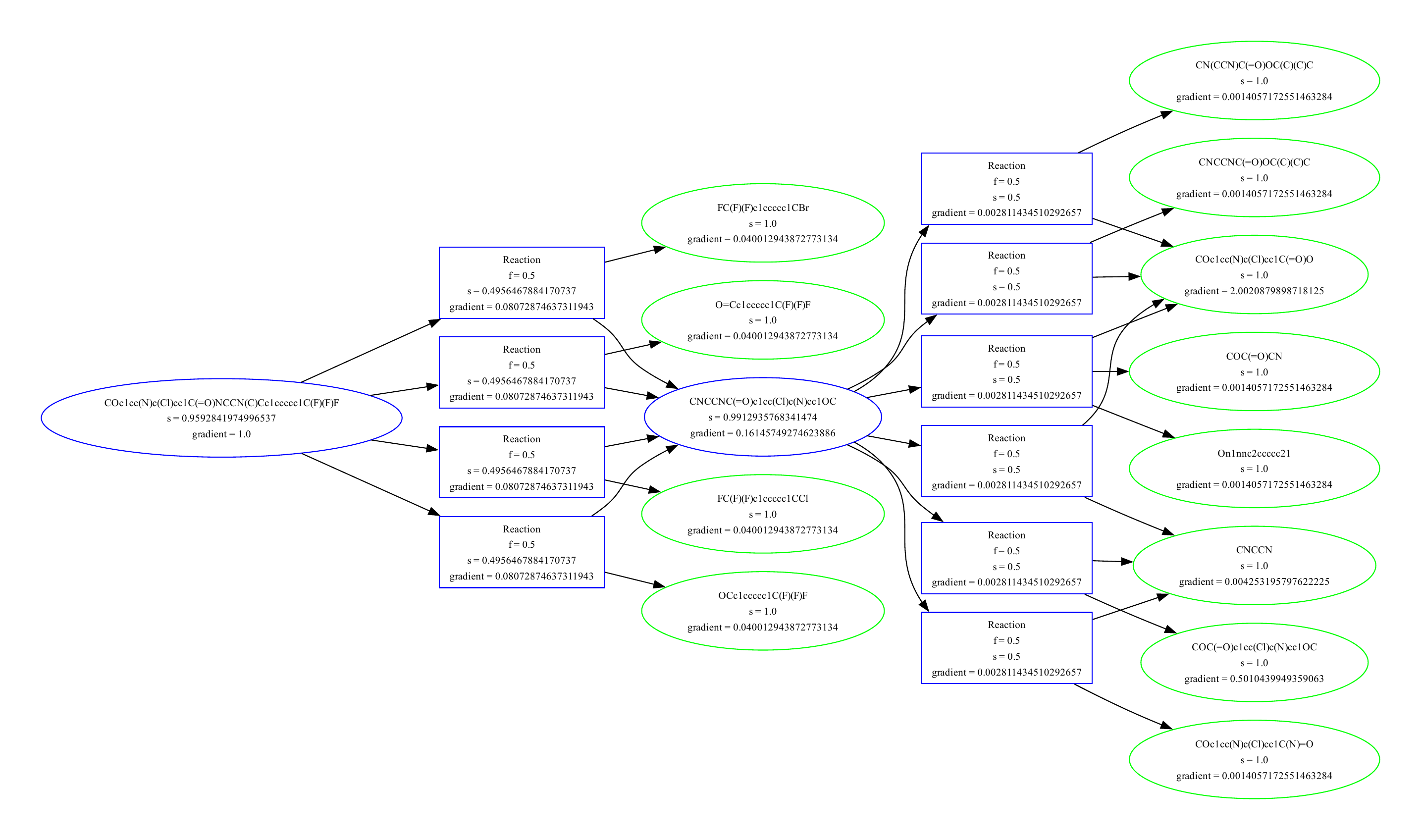}
\caption{The successful synthesis plan extracted from Figure~\ref{full_graph}.}
\end{figure}

\begin{figure}[ht]
\centering
\includegraphics[width=1.0\textwidth]{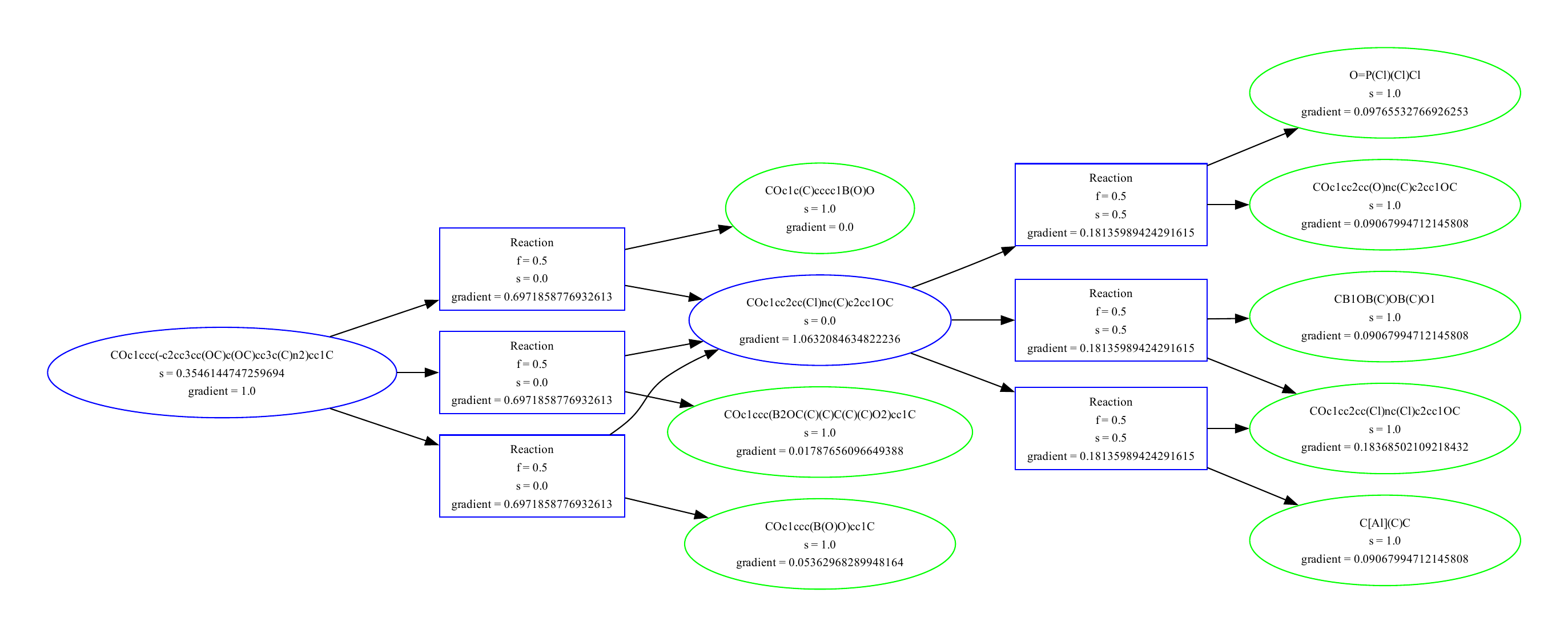}
\caption{The successful synthesis plan extracted from Figure~\ref{full_graph_2}.}
\end{figure}

\end{document}